\documentclass[two-columns]{nature} 
\usepackage{graphicx,caption} \usepackage{color} \usepackage{soul}
\usepackage{siunitx}
\usepackage{amsmath,physics}
\usepackage[unicode=true,bookmarks=true,colorlinks=true,
allcolors=blue]{hyperref} \usepackage{cleveref} \graphicspath{{./images/}}
\usepackage{textcomp} 
\usepackage{gensymb}
\usepackage{amssymb}
\usepackage[textsize=scriptsize]{todonotes}
\usepackage{latexsym}

\newcommand{\fast}[1]{\textcolor{green}{fast}}
\newcommand{\slow}[1]{\textcolor{orange}{slow}}

\newcommand{\Lwk}{\hat{\Lambda}_k}

\newcommand{\St}{\hat{\mathbf{S}}^{t}}
\newcommand{\Sij}{\hat{S}^{t}_{ij}}
\newcommand{\Sijk}{\hat{S}^{t}_{ijk}}
\newcommand{\B}{\boldsymbol{\beta}}
\newcommand{\Bt}{\boldsymbol{\beta}^{t}}
\newcommand{\Bij}{\beta^{t}_{ij}}
\newcommand{\Bijh}{\hat{\beta}^{t}_{ij}}
\newcommand{\Bth}{\hat{\boldsymbol{\beta}}^{t}}
\newcommand{\Ft}{\hat{\mathbf{F}}^{t}}
\newcommand{\Rt}{\hat{\mathbf{R}}^{t}}

\newcommand{\Es}{E_s}
\newcommand{\Em}{E_m}
\newcommand{\Dd}{D}

\usepackage[backend=biber,bibencoding=utf8, style=nature, url=false, isbn=false, doi=false,eprint=false]{biblatex}
\title{Artificial intelligence optical hardware empowers high-resolution hyperspectral video understanding at 1.2~Tb/s}

\author{M. Makarenko$^{1}$, A.
Burguete-Lopez$^{1}$, Q. Wang$^{1}$, S. Giancola$^{1}$, B. Ghanem$^{1}$, L. Passone$^{1}$,  
\& A. Fratalocchi$^{1*}$}

\begin{document}

\maketitle

\begin{abstract}
Foundation models, exemplified by GPT technology, are discovering new horizons in artificial intelligence by executing tasks beyond their designers' expectations. While the present generation provides fundamental advances in understanding language and images, the next frontier is video comprehension. Progress in this area must overcome the 1~Tb/s data rate demanded to grasp real-time multidimensional video information. This speed limit lies well beyond the capabilities of the existing generation of hardware, imposing a roadblock to further advances. This work introduces a hardware-accelerated integrated optoelectronic platform for multidimensional video understanding in real-time. The technology platform combines artificial intelligence hardware, processing information optically, with state-of-the-art machine vision networks, resulting in a data processing speed of 1.2~Tb/s with hundreds of frequency bands and megapixel spatial resolution at video rates. Such performance, validated in the AI tasks of video semantic segmentation and object understanding in indoor and aerial applications, surpasses the speed of the closest technologies with similar spectral resolution by three to four orders of magnitude. This platform opens up new avenues for research in real-time AI video understanding of multidimensional visual information, helping the empowerment of future human-machine interactions and cognitive processing developments.
\end{abstract}

\section*{Introduction}
\begin{figure*}
       \centering
	\includegraphics[width=.9\textwidth]{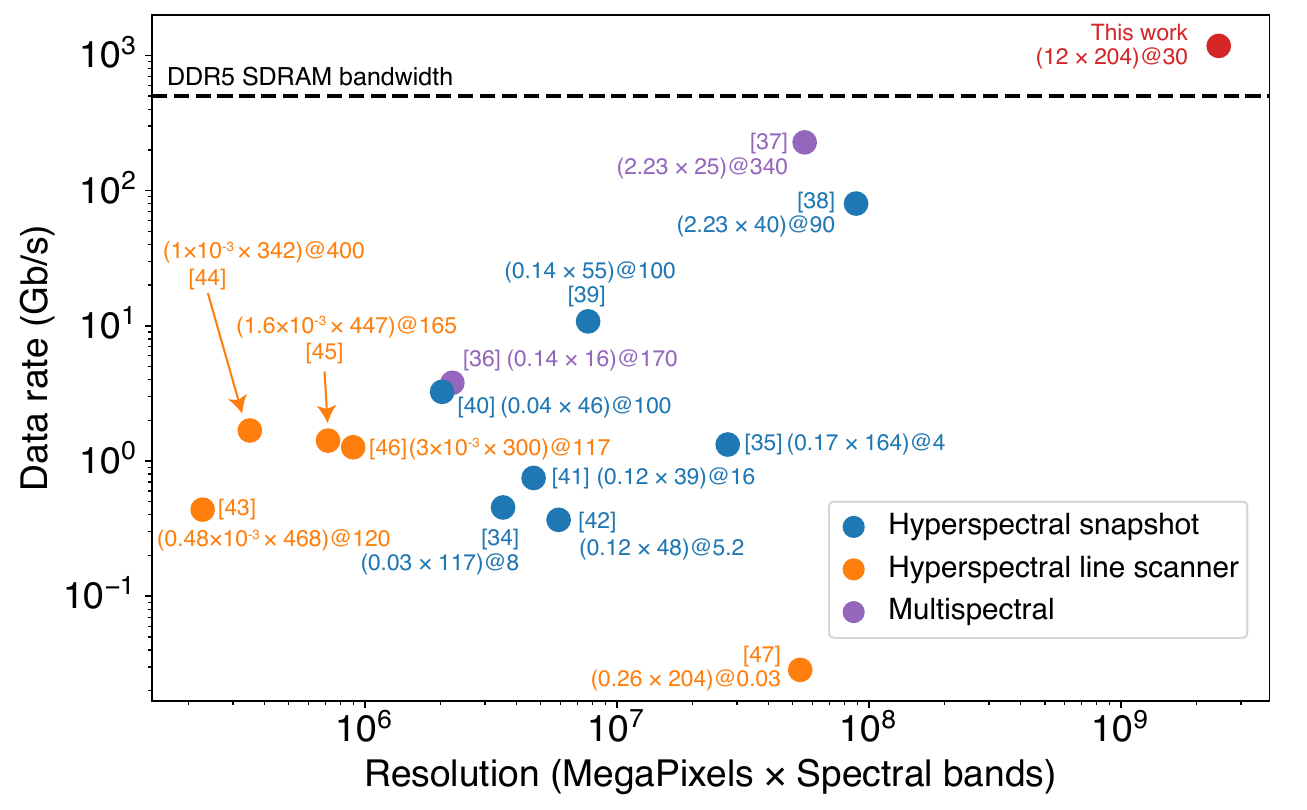}
	\caption{
		\label{fom} State-of-the-art in the acquisition of multidimensional optical information flow. Next to each technology, we indicate the number of megapixels times the number of spectral bands $@$ frames per second (FPS), as the producer specifies in the technology datasheet of the hardware used. The dashed line indicates a state-of-the-art DDR5 memory's ideal bandwidth, which provides the upper theoretical limit of any electronic technology requiring data transfer.
  }
\end{figure*}

The autonomous processing of big data via artificial intelligence (AI)~\cite{bigdata1, bigdata2, bigdata3} is opening frontiers in medical~\cite{Afromowitz1988,Panasyuk2007,lu2014medical,Gowen2015}, security~\cite{security1, security2}, robotics~\cite{robotics1, robotics2,image1,video1}, automated speech recognition~\cite{sequence}, and natural language processing~\cite{nlp} with human-like and ---in some cases--- better than human performances. Foundation models are accelerating this development significantly with the emergence of an understanding of human languages, carrying out tasks the designer never trained the model on~\cite{gpt4p,gpt3, gato}. The latest generation of foundation models in GPT-4 technology extend such learning abilities by combining information from languages and images using different data modalities~\cite{multimodal1, multimodal2,gpt4p, gpt4p}, while  
the recently proposed Gato model allows generalist agents to communicate, process pictures, play video games, and control robotic arms at the same time~\cite{gato, foundnature}.\\ 
The speed of these developments has significantly outpaced the velocity of hardware progress, with a roadblock looming on multimodal AI applications if new technological platforms to acquire and process data do not materialize~\cite{10.1063/5.0174044}. This issue is particularly significant in the emerging horizon of video understanding, representing the subsequent large language model (LLM) development, which requires grasping context in four-dimensional spatial and temporal data~\cite{Zhao_2023_CVPR,Buch_2022_CVPR,nips2023OmniVL}.\\
Hyperspectral imaging is the closest existing technology to multidimensional data flow acquisitions~\cite{Yu2014,Chen2018,lopez2023hyperspectral}. Hyperspectral imaging augments two-dimensional spatial pixels with a third dimension comprising hundreds of frequency channels corresponding to narrow portions of the spectrum within and beyond the visible range~\cite{lu2014medical}. These frequency bands encompass spectral signatures essential for the  identification, measurement, and classification of objects, materials, and compounds while enabling remote monitoring of their properties in diverse processes of industrial interest~\cite{moncrieff2002spectrophotometric,johansen2020recent,Afromowitz1988,Panasyuk2007,lu2014medical,Gowen2015,Chennu2017,HyperWater}. 
In the current standard for digital video~\cite{ITU-R_BT.2020}, a high-resolution hyperspectral datastream at 4K ($3840 \times 2160$ pixels), acquired with hundreds of bands in the visible range between the wavelengths of $350$~nm and $750$~nm, and with 12 bits per band, requires processing a data rate over \SI{1}{Tb/s}. Currently, the best snapshot hyperspectral devices capable of recording more than a hundred frequency bands possess between three to four orders of magnitude slower data rates, and cannot record at video speed~\cite{nguyen2013snapshot,cubert} (Fig.~\ref{fom}). Faster hyperspectral and multispectral technologies with frame rates in the \SI{100}{Gb/s} range reduce the spectral resolution by one order of magnitude~\cite{ximea,photonfocus,imec,dwight2018compact,pawlowski2019high,amann2023design,kester2011realtime}. At the same time, accurate one-dimensional scanners~\cite{corning,headwall,resonon,hyspex,specim} do not meet the spatial resolution required to capture 2D image flows at video rates. The critical challenge to reaching real-time Tb/s multimodal data processing is the speed of data transmission in electronics. State-of-the-art DDR5 memory, with a bandwidth of \SI{500}{Gb/s}~\cite{JDEC_DDR5}, exemplifies this barrier that the current technology cannot yet overcome (Fig.~\ref{fom} dashed line).\\
Addressing this problem provides a substantial opportunity in research to improve this technology significantly, unlocking future advancements in a vast number of critical applications in the medical, life sciences, forensics, security, pharmaceutical, environmental, mining, and oil industries that real-time cognitive processing of multidimensional visual data flows could empower~\cite{videoclass,videoseg,videotrack,Chennu2017,HyperWater,moncrieff2002spectrophotometric,johansen2020recent,Afromowitz1988,Panasyuk2007,lu2014medical,Gowen2015,menguSnapshotMultispectral2023,liSpectrallyEncoded2021}.

\section*{Results}
\begin{figure*}
       \centering
	\includegraphics[width=\textwidth]{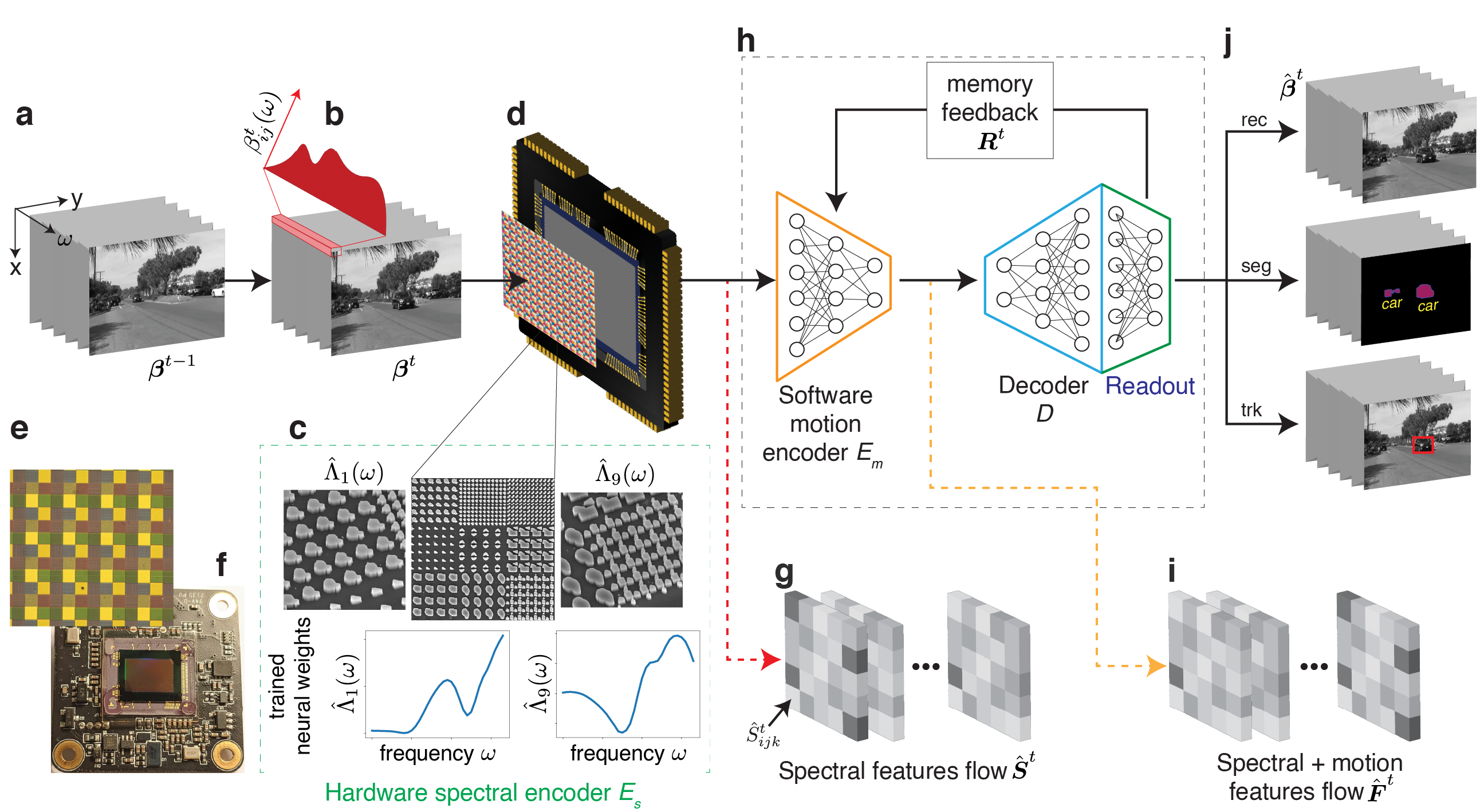}
	\caption{
		\label{concept}	\textbf{Hardware-accelerated video understanding platform.} (a) An example of hyperspectral video flow comprising frames $\Bt$ over time. (b) Power density distribution $\beta_{ij}^t(\omega)$ emanating from one pixel $i,j$ in the scene (c) Scanning electron microscope (SEM) images of the hardware encoder $E_s$ composed of a set of nanoresonators with trained transmission functions $\Lambda_k(\omega)$ acting as neural weights for feature extraction. (d) Schematics of hardware encoder placed on top of a camera sensor. (e) Microscope image of a fabricated $N_k=9$ encoder array. (f) The visual appearance of an experimental board camera sensor integrated with the hardware encoder of (e). (g) Schematic illustration of spectral features flow read by (d). (h) Recurrent AI module comprising a motion encoder, decoder unit, and readout. (i) Schematic representation of combined spectral and motion features. (j) Different video tasks computed by this platform.}
\end{figure*}

Figure~\ref{concept} illustrates the architecture of a hardware-accelerated platform for real-time hyperspectral video understanding we propose to address the aforementioned issues. A video sequence comprising a succession of frames $(\dots,\B^{t-1},\B^t,\dots)$ with $\B^{t}$ corresponding to the frame at time $t$ represents the input of the system (Figs.~\ref{concept}a-b). Each frame $\B^{t}$ contains a three-dimensional data representation of the optical information flow, with $\Bij(\omega)$ representing the power density spectrum emanating from a single spatial point $(i,j)$ in the scene (Fig.~\ref{concept}b, solid red area).\\
A hardware encoder $\Es$ (Fig.~\ref{concept}c) extracts spectral features from each video frame $\B^{t}$. The hardware encoder comprises an array of nanostructured pixel encoders positioned on top of a camera sensor (Fig.~\ref{concept}d). Each pixel contains a set of $k=1,...,N_k$ nanoresonator configurations with transmission functions $\hat\Lambda_k$ (Fig. \ref{concept}c). Figures~\ref{concept}e-f shows the physical implementation of this concept, presenting the pixel array (Fig.~\ref{concept}e) integrated on a monochrome camera sensor board (Fig.~\ref{concept}f). Fig.~\ref{concept}e shows a view of the encoder pixel array at $100\times$ magnification, with each colored square corresponding to a different encoder.\\
When the input data flow impinges on one pixel, the camera sensor converts the spectrum emanating from a spatial point of the video frame into a digital scalar coefficient $\Sijk$ read by the camera hardware:
\begin{equation}
\label{mac0}
    \Sijk = \sigma\left(\int \Bij(\omega) \Lwk(\omega) \dd{\omega}\right),
\end{equation}
where $\sigma(x)$ is the readout input-output response of the single camera pixel~\cite{contires}, or an added nonlinearity implemented in software, and $\Lwk(\omega)$ is the encoder pixel's transmission function. We inversely design the nanoresonators in each pixel using universal approximators as described in~\cite{Makarenko2021robust, Getman2021, Makarenko_2022_CVPR}. Because these nanostructures can approximate arbitrary responses~\cite{Getman2021}, it is possible to inverse-design the transmission $\Lwk(\omega)$ of each pixel encoder to represent any user-defined distribution of amplitude coefficients for the spectral coordinate $\omega$. In this condition, the camera integration \eqref{mac0} implements the hardware equivalent of a neural network's multiply-accumulate (MAC) operation, where $\Bij(\omega)$ represents the input and $\Lwk(\omega)$ represent neural weights distributed along the frequency axis $\omega$. We train the weights $\Lwk(\omega)$ to implement feature extraction tasks of a traditional software neural network (see Methods). 
The hardware encoder performs this operation at optical speed and in parallel for every pixel, generating a flow of sparse spectral features $\St$ (Fig.~\ref{concept}g).
The camera hardware reads the features flow $\St$ and sends it to the software motion encoder $\Em$ (Fig.~\ref{concept}h).
The encoder $E_m$ combines spectral and motion features extracted from the data flow into the feature flow tensor $\Ft$ (Fig. \ref{concept}i). The motion features comprise dynamic temporal changes between video frames, including the direction and speed of movement of objects, changes in the flow composition, and variations in illumination over time.
The motion encoder processes these features in real-time with a memory feedback $\Rt$ comprising information extracted from previous time-frames (Fig.~\ref{concept}h, feedback loop). The feature flow projects sequentially into a decoder $\Dd$
terminating in a nonlinear readout (Fig.~\ref{concept}h, right side). The decoder processes the feature tensor $\Ft$ for different end-to-end optimizations of user-defined tasks, including spectral video reconstruction $\hat{\boldsymbol{\beta}}^t$, video object segmentation~\cite{videoseg}, and spectral object tracking~\cite{videotrack} (Fig.~\ref{concept}j).

\subsection{Spectral video reconstruction}
\begin{figure*}
\centering
\includegraphics[width=.9\textwidth]{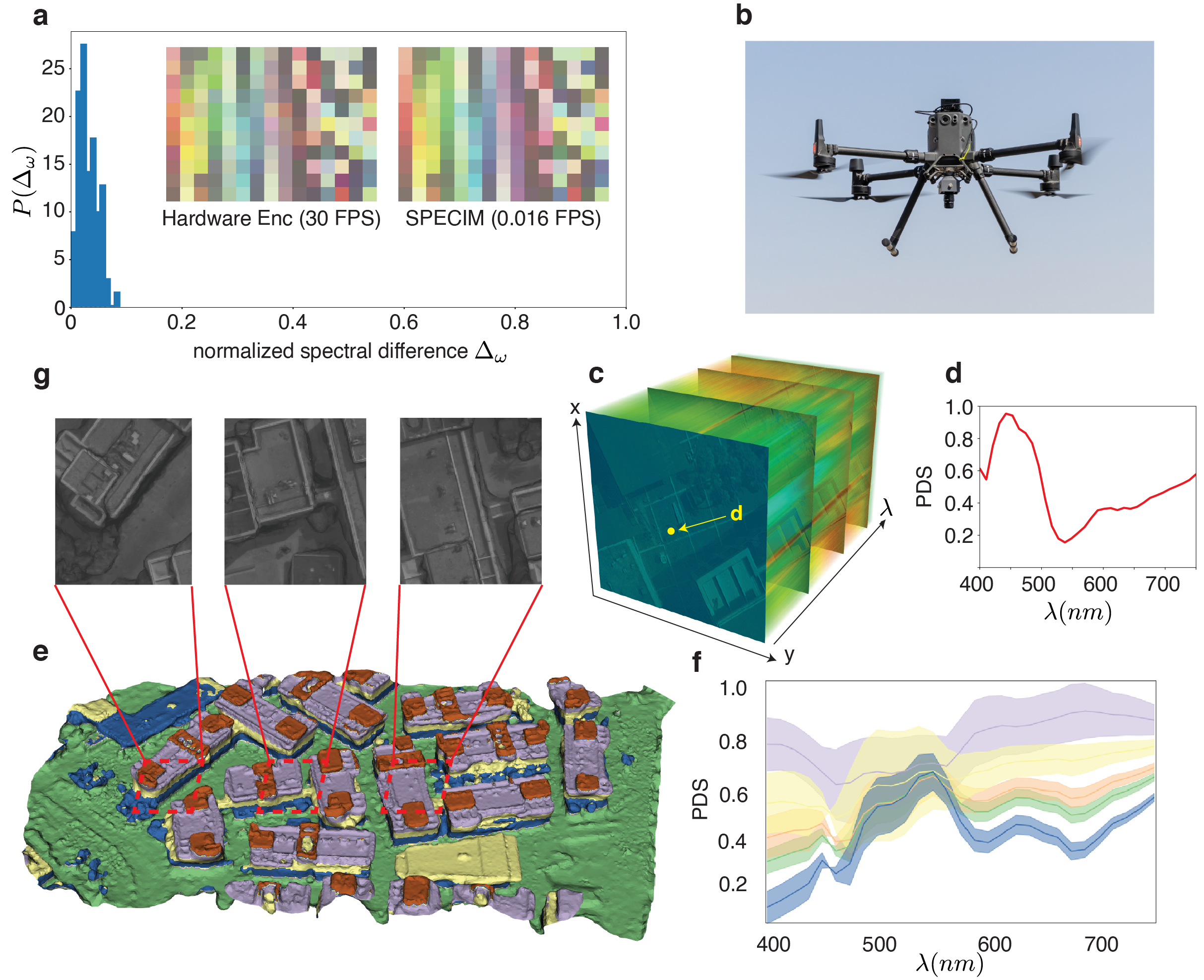}
\caption{
    \label{reconstruction}	\textbf{Real-time hyperspectral UAV scene reconstruction.} (a) Distribution of spectral reconstruction difference for a reconstructed hyperspectral frame image (inset) at 204~bands using the platform of Fig.~\ref{concept} at 30 FPS and a commercial SPECIM IQ operating at 0.016~FPS. (b) Camera module with hardware encoders integrated on DJI Matrice 300 RTK drone. (c) Visualization of a single hyperspectral frame acquired by the drone at 30~FPS. (d) Power density spectrum (PDS) retrieved at the spatial pixel (d) in panel c. (e) Reconstructed three-dimensional hyperspectral map from the video sequences, with spectral distribution visualized via K-MEANS clustering. (f) Cluster's spectral distribution. (g) Raw frames in the hyperspectral UAV video sequence representing selected areas in panel e.}
    \end{figure*}

The goal of spectral reconstruction is predicting the visual flow $\Bth$ while minimizing the difference $\Delta_\omega=||\Bt-\Bth||$ with the original input $\Bt$ (Fig.~\ref{concept}j, rec). We configure the system of Fig.~\ref{concept} for this task by disconnecting the recurrent feedback $\Rt$ unit (Fig.~\ref{concept}h, top) and set the motion encoder $\Em$ as the identity operator $I$, carrying out all encoding operations through the hardware encoder $\Es$. 
For the type of encoding used in this work (see Methods), we decode the information flow through the projector $\Dd=\boldsymbol{\hat{\Lambda}}^\dag(\omega)$, with $\boldsymbol{\hat{\Lambda}}(\omega)=\qty[\hat{\Lambda}_1(\omega),...,\tilde{\Lambda}_k(\omega)]$ being the set of trained encoder transmission functions. When the projector $D$ operates on the features flow $\Sij$ arising from one camera pixel $i,j$, we obtain $\Bijh(\omega)=D\Sij=\boldsymbol{\hat{\Lambda}}^\dag(\omega) \Sij=\sum_k \hat{\Lambda}_k(\omega)\Sijk$, which represents the optimal least square approximation of the spectral video information flow~\cite{jolliffe2016principal}. In this computation we perform a demosaicing process analogous to the Bayer array interpolation step of RBG cameras, which allows retrieving the spectral features flow at every camera pixel, maintaining the resolution of the camera sensor regardless of the number $N_k$ of encoders used.\\
Figure~\ref{reconstruction} presents experimental results of the hardware-accelerated hyperspectral platform in field applications. Figure~\ref{reconstruction}a validates the reconstruction of a single video frame by using a calibration palette comprised of various colors with known reflection spectra. The test compares two hyperspectral images with 204 frequency bands each, one captured with our hardware-accelerated camera with 12 Megapixels, $N_k=9$ trained encoders on a publicly available general hyperspectral dataset~\cite{wolfgang}, and a video rate of 30~FPS, and the other obtained with a commercial SPECIM IQ (Specim, Spectral Imaging Ltd.) hyperspectral camera, which possesses 0.26 Megapixels and a 0.016 FPS acquisition rate. The solid area in Fig.~\ref{reconstruction}a shows the distribution $P(\Delta_\omega)$ of the absolute spectral difference $\Delta_\omega$ between the data retrieved with the two cameras. The hardware accelerated platform, while working at an acquisition rate approximately 2000 times faster, provides the same spectral prediction for the same number of bands, with an average reconstruction difference below 3\%. The insets in Fig.~\ref{reconstruction}a show reconstructed RGB images of the palette from hyperspectral data.\\
Figures~\ref{reconstruction}b-g illustrate a field application for real-time video understanding using an Unmanned Aerial Vehicle (UAV). We integrate the hardware accelerated platform of Fig.~\ref{concept}f on a Matrice 300 RTK drone ~(Fig.~\ref{reconstruction}b), and then record a total of \SI{80}{\minute} of aerial hyperspectral video footage at 30 frames per second, capturing the terrain from an altitude of \SI{50}{m}. Figure~\ref{reconstruction}c shows a single hyperspectral video frame, while Fig.~\ref{reconstruction}d illustrates the reflection spectra associated with one spatial point (panel c, yellow dot). Figure~\ref{reconstruction}e shows a complete 3D hyperspectral map representing a \SI{0.5}{km^2} area of the terrain obtained by processing the aerial footage. We compute the map using Pix4D Mapper~\cite{PIX4DmapperProfessional}, a photogrammetry software that stitches video frames into 3D models of surveyed locations. We implement a K-means clustering algorithm to analyze the hyperspectral data further, segmenting it into four distinct clusters based on depth and spectral dimensions~(Fig.~\ref{reconstruction}f). We visualize these clusters by plotting each pixel of the 3D hyperspectral map with the corresponding spectral color in panel f. The red pixel cluster in the map represents high-elevation areas of the buildings hosting metallic heating, ventilation, and air conditioning (HVAC) units. Light violet pixels correspond to rooftop surfaces, which the clustering algorithm segments as the strongest light-reflecting scene objects. Yellow pixel clusters map concrete, which appears as the buildings' walls and the building's rooftop in the lower right area where dust has accumulated. The blue pixels combine information about objects above the ground that predominantly reflect green light, marking the signature of chlorophyll and indicating specific species of vegetation. Finally, green pixels map the ground plane. Figure~\ref{reconstruction}g displays a selection of reconstructed frames from the original hyperspectral footage~(Fig.~\ref{reconstruction}e, dashed lines).

\subsection{Hyperspectral video segmentation and tracking.}
\begin{figure*}
       \centering
	\includegraphics[width=.7\textwidth]{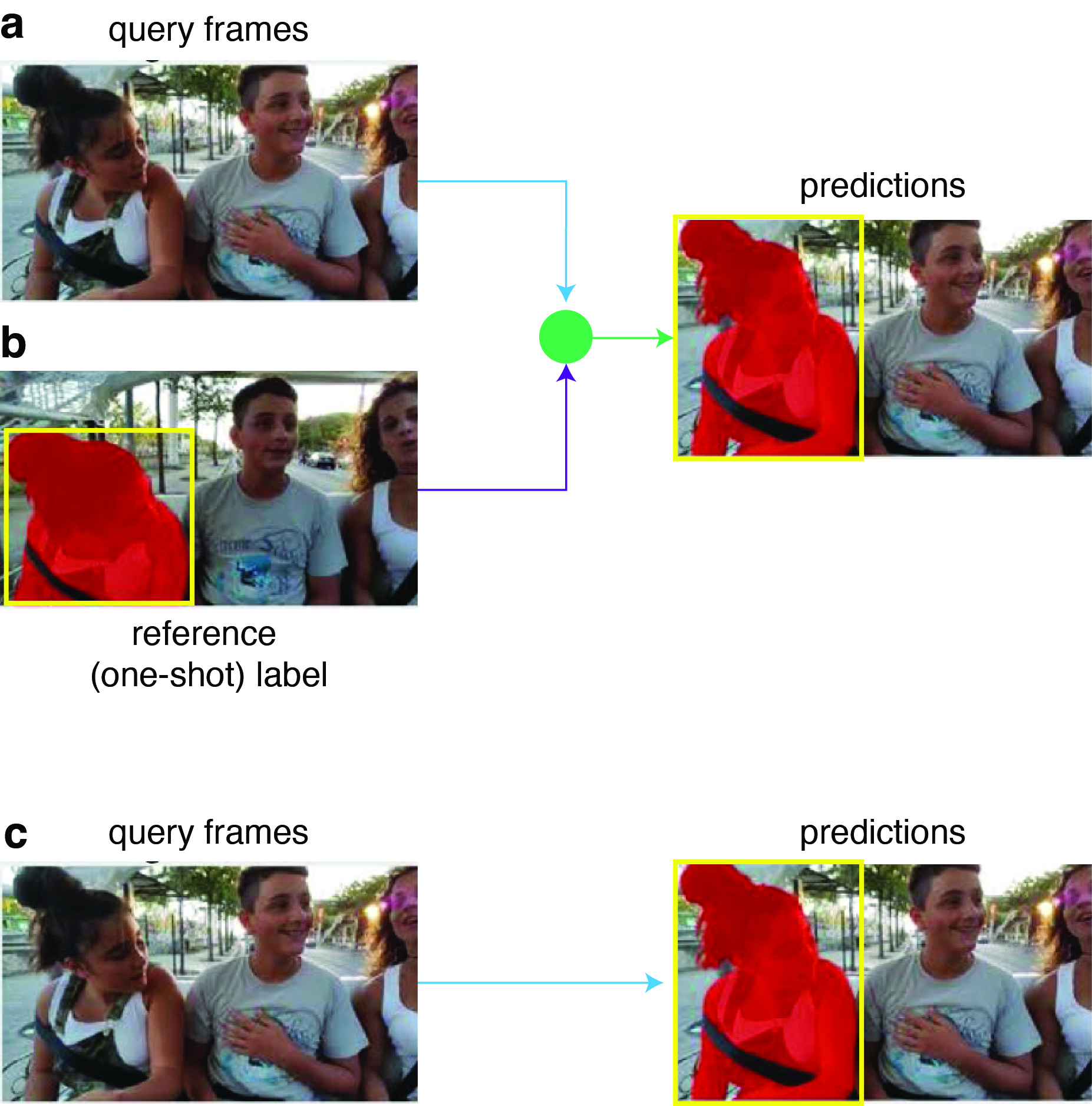}
	\caption{
		\label{vos} \textbf{Approaches to AI video segmentation and tracking:} (a-b) One-shot video object segmentation (OVOS) and (c) zero-shot video object segmentation (ZVOS).}
\end{figure*}

\begin{figure*}
       \centering
	\includegraphics[width=.99\textwidth]{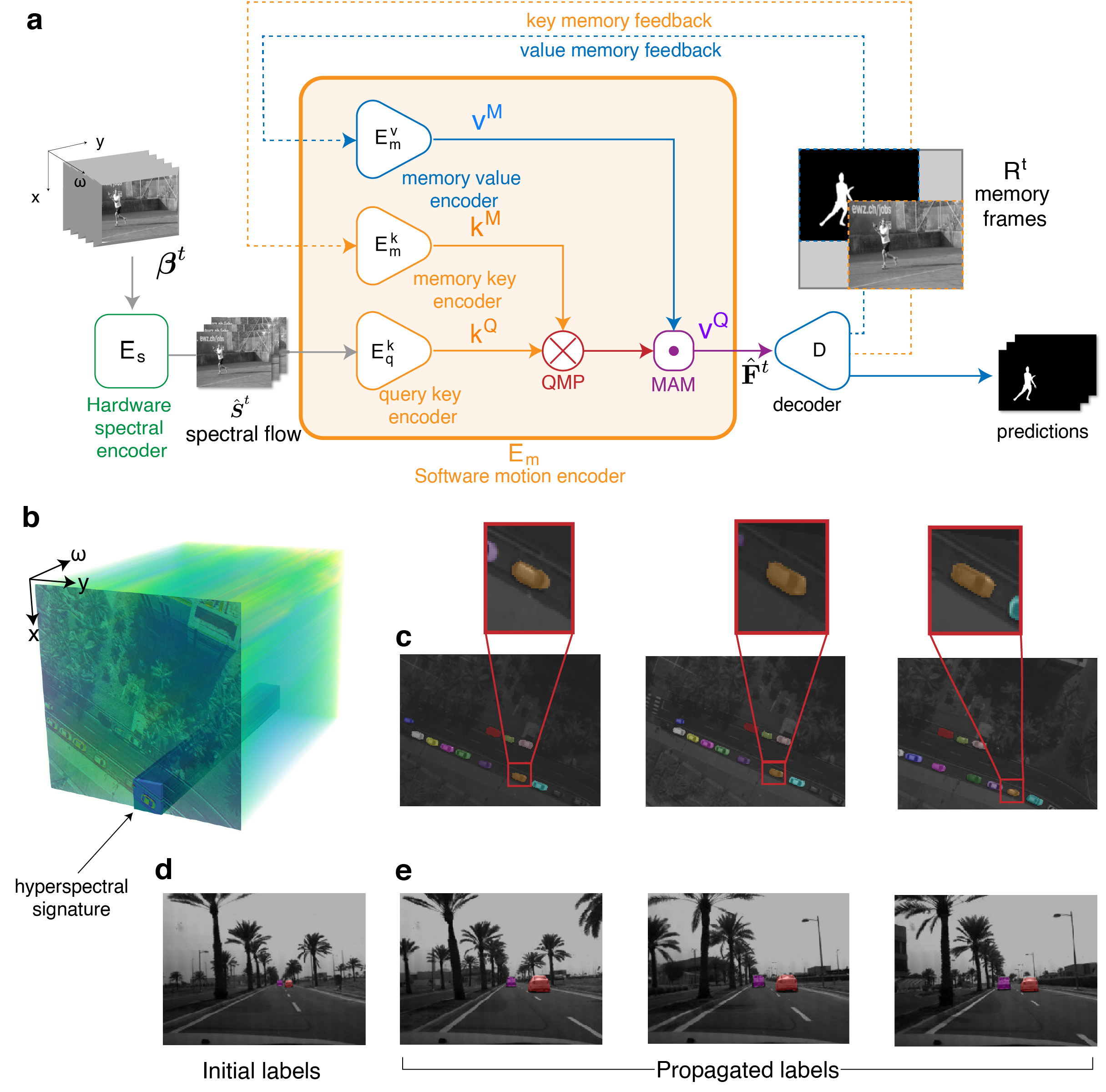}
	\caption{
		\label{tracking} \textbf{One-shot hyperspectral segmentation and object tracking.} (a) Configuration of the general architecture of Fig.~\ref{concept} for one-shot video semantic segmentation. (b-c) Segmentation results on hyperspectral object tracking with (b) Hyperspectral signature selected from the datacube acquired by the UAV of Fig.~\ref{reconstruction}. (c) Segmentation results in real-time with detailed zoomed regions. (d) Video recording from inside a car and initially selected signatures. (e) A segment of hyperspectral data with unique signatures that the architecture tracks and propagates over time. }
\end{figure*}

\begin{figure*}
       \centering
	\includegraphics[width=.88\textwidth]{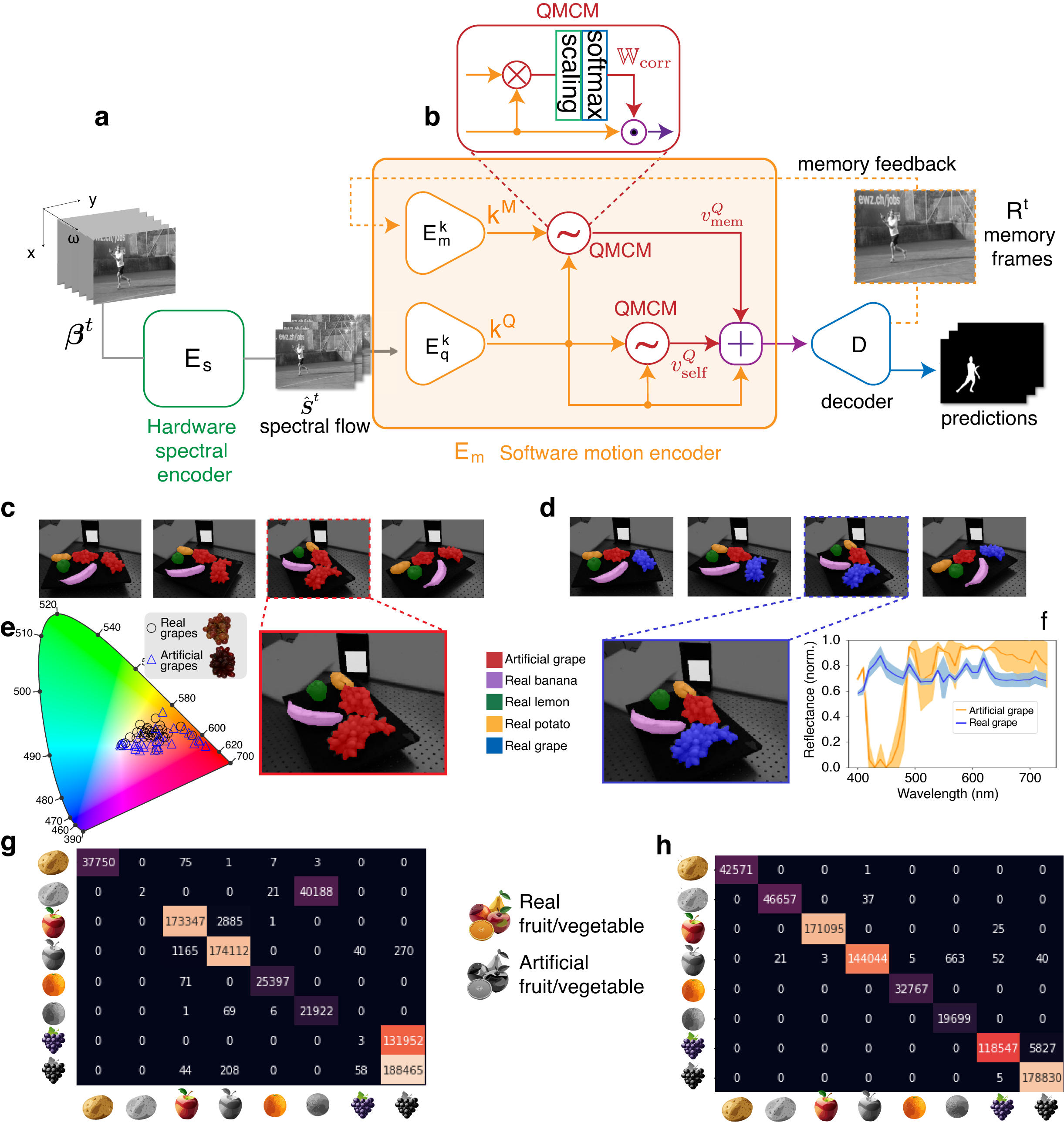}
	\caption{
		\label{segmentation} \textbf{Zero-shot object segmentation and tracking} (a) Hardware and software AI architecture scheme for ZVOS. (b) The inner architecture of the query memory correlation block. (c-d) Semantic segmentation resulting from (c) RGB and (d) hyperspectral data. (e) Natural (cross markers) and artificial (plus markers) grape chromaticity visualized in the CIE 1931 color space. (f) Hyperspectral data of (solid yellow lines) artificial and (solid blue line) natural grapes. (g-h) Confusion matrix on the model on (g) RGB and (h) hyperspectral data.}
\end{figure*}

Video Object Segmentation (VOS) in AI video understanding aims to classify and monitor target objects distinct from the background across video frames over time. Figure~\ref{vos} illustrates the two principal methodologies in VOS. One-Shot Video Object Segmentation (OVOS)~\cite{segmentation_review} uses manually labeled reference frames to instruct the segmentation algorithm on the initial composition of targets (Figs.~\ref{vos}a-b). This technique is semi-supervised and necessitates human input to specify the objects of interest. Target labeling uses either a segmentation map (Fig.~\ref{vos}b, red transparency) or a bounding box for visual object tracking (Fig.~\ref{vos}b, yellow rectangle). Zero-Shot Video Object Segmentation (ZVOS), conversely, autonomously processes objects with different visual characteristics without human-defined labels (Fig.~\ref{vos}c).\\ 
The main limitation of current VOS processes is that AI cannot segment information that cameras cannot capture, that is, light beyond RGB colors. The hardware-accelerated hyperspectral platform introduced in this work addresses this problem by empowering AI with spectral features, which provide more comprehensive information than primary colors.
Figure~\ref{tracking}a illustrates the configuration of the hardware-accelerated video understanding platform for one-shot video semantic segmentation. In this system, the spectral flow arising from the hardware encoder enters a motion encoder $\Em$ comprising five integral modules that incorporate state-of-the-art spatial-time network models~\cite{stn, stcn, perclip}. The first module, the query key encoder (Fig.~\ref{tracking}a, $E_q^k$ unit), extracts spectral-spatial image features $k^Q$, which the query-memory projection (Fig.~\ref{tracking}a, QMP) processes. The QMP computes similarities between the $k^Q$ flow and spectral-spatial features $k^M$ arising from previous frames extracted by the memory key encoder (Fig.~\ref{tracking}a, $E_m^k$ unit). The QMP evaluates the degree of affinity via a similarity matrix $\mathbb{W} \in \mathbb{R}^{H W \times H W}$:
\begin{equation}
\mathbb{W}\left(k^Q, k^M\right)_{i, j}=\frac{\exp \left(k_i^Q \odot k_j^M\right)}{\sum_j \exp \left(k_i^Q \odot k_j^M\right)},
\end{equation}
where the $\odot$ operator is the dot product. The matrix entries $\mathbb{W}_{i, j}$ furnish a similarity score between the input flows $k^Q$ and $k^M$ ranging between zero and one. The mask adjustment module (Fig.~\ref{tracking}a, MAM)  processes the data by projecting the video mask $v^M$, computed from previous frames by the memory value encoder (Fig.~\ref{tracking}a, $E_m^v$ unit), to the QMP output using the following similarity matrix $\mathbb{W}$:
\[
v^Q=\mathbb{W}\left(k^Q, k^M\right) v^M.  
\]  
The mask adjustment module output represents the feature tensor $\Ft$ for the current timestep $t$ (Fig.~\ref{concept}i). The output decoder further processes $\Ft$ to predict the mask for the frame at timestep $t$, and output the frame itself. The mask and frame then form feedback for the motion encoder's successive predictions.\\
Figures~\ref{tracking}b-e presents the results on OVOS using the hardware-accelerated platform of this work operating with 204 bands (OVOS204) applied to two distinct segmentation tasks. The first task (Figs.~\ref{tracking}b-c) uses hyperspectral video data acquired with the UAV of Fig.~\ref{reconstruction} to track and segment the hyperspectral signature characterizing a specific car from many with the same visual color appearance in the aerial footage (Fig.~\ref{tracking}b). Accessing unique hyperspectral signatures allows the user to find and correctly label objects that color cameras cannot distinguish due to a lack of information. Figure~\ref{tracking}c shows how the system successfully tracks the target's position over time. The images in Fig.~\ref{tracking}c depict the data evolution directly from the spectral flow datastream received by the camera, which contains the features motion encoder will process. In the second task (Fig.~\ref{tracking}d), we mounted the hardware accelerated camera inside a car and performed the same type of hyperspectral tracking, segmenting spectral signatures corresponding to specific vehicles (Figs.~\ref{tracking}d-e).\\ 
Figure~\ref{segmentation} showcases additional ZVOS examples of how hyperspectral video flows empower AI understanding. Figures~\ref{segmentation}a-b show the system configuration for ZVOS. A query encoder module (Fig.~\ref{segmentation}a, $E^k_q$ unit) extracts spectral-spatial features $k^Q$ using the same encoder architectures as in the OVOS task. A query memory correlation module processes these features nonlinearly (Figs.~\ref{segmentation}a-b, QMCM). The QMCM understands dense spatial relationships in the input frame features by using the correlation matrix $\mathbb{W}_{\text{corr}}$, defined as: \[ \mathbb{W}_{\text{corr}}(k^M, k^Q) = \frac{1}{C_k}\text{softmax}\qty[k^M (k^{Q})^{\dag}]. \] to project the query key feature vector \( k^Q \):
    \[ v_{\text{mem}}^Q = \mathbb{W}_{\text{corr}}(k^M, k^Q) k^Q. \]
The top QMCM understands relationships between present $k^Q$, and past $k^M$ query features arising from the memory encoder $E^k_m$. The bottom QMCM understands the correspondence of $k^Q$ features with themselves. The motion encoder $E_m$ concatenates the output from both QMCMs with the initial flow $k^Q$. The decoder uses the same architecture as in the OVOS scheme, outputting the predicted mask alongside the frame, which becomes feedback for $E_m$ as a memory frame $R^t$.\\
We benchmark the hardware-accelerated platform on a hyperspectral video dataset (FVgNET-video) built from the public FVgNET dataset~\cite{Makarenko_2022_CVPR}. FVgNET-video consists of \SI{30}{FPS} hyperspectral video sequences of combinations of artificial and natural fruits and vegetables placed on a rotating turntable (Figs.~\ref{segmentation}c-d). Figures.~\ref{segmentation}c-h illustrate the performance of hardware-accelerated hyperspectral ZVOS on samples from the dataset. We compare the performance of the hyperspectral camera with a simulated RGB camera that records at the same resolution and framerate. Figures~\ref{segmentation}c-d show segmentation maks generated in real-time from a video sequence showing two grapes, one banana, one orange, and one potato on the turntable.
One of the grape bundles is artificial, while the rest are natural. Figure~\ref{segmentation}c shows the result of segmentation masks created on RGB data, with each color marking a different object class. These images show that the RGB camera cannot distinguish between artificial and natural objects, predicting that both grapes are of the same type. Figure~\ref{segmentation}d shows the segmentation masks resulting from hyperspectral data, allowing AI to identify all items correctly. Figure~\ref{segmentation}e presents the CIE 1931 chromaticity diagram distribution of the RGB values for the artificial and natural grape bundles of the video sequence. The panel shows there is little chromaticity variation between these samples, causing the RGB VOS to fail. Figure~\ref{segmentation}d shows the reflection spectra for each grape bundle. In contrast to panel e, there is significant variation in the spectral response of the two objects, explaining the success of the hyperspectral VOS. Figures~\ref{segmentation}g-h quantify this performance difference further by showing the confusion matrices for the RGB and hyperspectral VOS tasks respectively. In the RGB case, the segmentation fails in two of the eight categories, with a substantial number of incorrect pixels for artificial oranges and real grapes. There is also significant confusion in the case of artificial grapes, with over 40\% of all pixels classified incorrectly. Conversely, the segmentation is successful for all categories in the hyperspectral case, with less than 5\% of the pixels incorrectly classified.

\section*{Conclusion}
This work implemented and field-validated a hardware-accelerated platform for real-time hyperspectral video understanding, demonstrating hyperspectral UAV scene reconstruction, hyperspectral video object segmentation, and classification using more than $200$ frequency bands, a 12 megapixel spatial resolution, and video rates of 30 FPS. This platform technology processes information beyond 1~Tb/s, enabling the current generation of AI to understand information that color video acquisition systems do not discern, and current hyperspectral imaging technologies cannot acquire in real-time and at these resolutions.  
This work opens new research and application opportunities for AI video understanding utilizing broadband hyperspectral data flows for environmental monitoring, security, pharmaceutical, mining, and medical diagnostics that require processing high-resolution spectral and spatial information at video rates. 
Future research can focus on developing scalable systems that benefit both the adoption of this technology and the ease of acquiring real-world hyperspectral data flows for subsequent AI development. The principles and methodologies devised in this work can also be generalized across various fields and help impact the way AI interacts with the visual world, particularly for foundation models like GPT-4 and Claude2. The technology we have introduced could empower this generation's AI to apply its information processing capacities to a broader set of multimodal tasks, facilitating the pursuit of more robust forms of AI such as Artificial General Intelligence~\cite{Shevlin2019-xi}. 

\section*{Methods}
\subsection{Hardware encoder nanofabrication.}
We employ a \SI{15}{mm} wide and \SI{500}{\micro m} thick square piece of fused silica (University Wafer) as our substrate. On it, we grow a layer of hydrogenated amorphous silicon (a-Si:H) using plasma enhanced chemical vapor deposition (PECVD). We then spin coat electron beam resist and print the nanostructure pattern using a JEOL JBX-6300FS electron beam lithography system. Following this, we develop the resist and perform a liftoff process, creating a hard mask with the shape of the nanostructures on the silicon. Finally, we use reactive ion etching (RIE) to completely remove the unprotected silicon, and remove the hard mask.

\subsection{Hardware encoder training.}
It is possible to train the encoders response functions $\Lwk(\omega)$ 
using both linear and nonlinear feature extraction schemes. In both cases, the starting point is to flatten the hyperspectral tensor into a single matrix containing the power density spectra of a set of camera pixels on each column, creating a dataset $\B$ for subsequent training. To provide feature extractions to this data, we can use any supervised or unsupervised techniques developed in deep learning to train the relevant distributions of coefficient $\Lambda(\omega)$. Once these distributions of values are found, we use ALFRED~\cite{Makarenko2021robust, Getman2021}, an advanced optimization framework for the design and implementation of nanoresonators with user-defined broadband responses across the visible and infrared. This article uses a linear encoder $\boldsymbol{\Lambda}$ developed through an unsupervised learning approach utilizing Principal Component Analysis (PCA). This process entails hardware encoding $\Es$, in which we specifically selected the nine strongest principal components, denoted as $\boldsymbol{\hat{\Lambda}^\dag}$, following the singular value decomposition of data tensor $\B$.\\
While it is possible to train the hardware encoder to the specific problem, we here use in every application the same set of encoders trained from a publicly available dataset of general hyperspectral images under various illumination conditions \cite{wolfgang}.

\printbibliography
\newpage

\clearpage

\end{document}